\documentclass[runningheads]{llncs}
\usepackage{graphicx}
\usepackage{amsmath}
\usepackage{amsfonts}
\usepackage[ruled]{algorithm2e} 
\usepackage{color}
\usepackage{multirow}
\usepackage{subfigure}

\usepackage[misc]{ifsym}

\begin{document}
\title{Model Bridging: Connection between Simulation Model and Neural Network}
\toctitle{Model Bridging: Connection between Simulation Model and Neural Network}
%
%
\author{Keiichi Kisamori \inst{1,2} \and
Keisuke Yamazaki\inst{2} \and
Yuto Komori\inst{2} \and
Hiroshi Tokieda\inst{2}}
\tocauthor{Keiichi Kisamori}
\authorrunning{K. Kisamori et al.}
%
\institute{NEC Corporation, Kanagawa, Japan \and 
National Institute of National Institute of Advanced Industrial Science and Technology, Tokyo, Japan}
\maketitle              
\begin{abstract}
The interpretability of machine learning, particularly for deep neural networks, is crucial for decision making in real-world applications.
One approach is replacing the un-interpretable machine learning model with \emph{a surrogate model}, which has a simple structure for interpretation.
Another approach is understanding the target system by using a simulation modeled by human knowledge with interpretable simulation parameters.
Recently, simulator calibration has been developed based on kernel mean embedding to estimate the simulation parameters as posterior distributions.
Our idea is to use a simulation model as an interpretable surrogate model.
However, the computational cost of simulator calibration is high owing to the complexity of the simulation model.
Thus, we propose a ``model-bridging'' framework to bridge machine learning models with simulation models by a series of kernel mean embeddings to address these difficulties.
The proposed framework enables us to obtain predictions and interpretable simulation parameters simultaneously without the computationally expensive calculations of the simulations.
In this study, we apply the proposed framework to essential simulations in the manufacturing industry, such as production simulation and fluid dynamics simulation.

\end{abstract}
%
%
%

\section{Introduction}

The interpretability of machine learning, especially for deep neural networks, is crucial for decision making in real-world applications.
In recent years, many studies have addressed the interpretability of neural networks~\cite{Guidotti2018,DoshiVelez2017,Molnar2019}.
One of the approaches is replacing the un-interpretable machine learning model with \emph{a surrogate model}, which has a simple structure for interpretation.
This approach is a type of model compression.
For instance, the ``distillation'' of a neural network model~\cite{Hinton2015} is one of the representative methods for model compression for replacing a complex model with a simplified model; however, there is no interpretability for a small surrogate neural network model.
There are some methods to obtain an interpretable model, such as LIME~\cite{Ribeiro2016}, SHAP~\cite{Lundberg2017}, and a method combined with a rule-based model~\cite{Wang2019}.
These methods do not provide a clear pathway toward obtaining the interpretability of a neural network, as there are limitations to obtain local interpretability regarding the decision boundary of the prediction result~\cite{Guidotti2018,DoshiVelez2017}.

Another approach for understanding the target system is by employing a simulation that might be outside the scope of conventional machine learning.
In some application fields, simulations such as multi-agent simulation, traffic simulation, production simulation, or simulation of the dynamics of the physical system have already been used to understand the target system and to predict future behavior.
Simulation modeling is implemented to describe the fundamental law of the objective system, using human knowledge with interpretable simulation parameters.
The recently developed ``simulator calibration''~\cite{Kennedy2001,Kisamori2018,Cleary2020} is a method in which the simulation parameters are estimated as posterior distributions in the context of machine learning.
Simulator calibration can provide a predictive result with interpretable simulation parameters.
Our idea is to use a simulation model as an interpretable surrogate model.
However, the difficulty of simulator calibration is attributed to a substantial computational cost; it typically takes more than one hour owing to the complexity of the simulation model (Table~\ref{table:comparison_ml_sim}).
In real-world applications, a predictive result and its reason often should be required to obtain within a minute.

\begin{table}[t]
\centering
\caption{Comparison between machine learning models and simulation models.}
\begin{tabular}{c|cc}
 & machine learning model \ \ \ & simulation model \\ \hline
interpretability of parameter & un-interpretable & interpretable \\
 computational cost of the model & not expensive & expensive \\
\end{tabular}
\label{table:comparison_ml_sim}
\end{table}

\begin{figure}[t]
\centering
\vspace{-0.1cm}
\includegraphics[width=9.cm, bb=0 0 604 233]{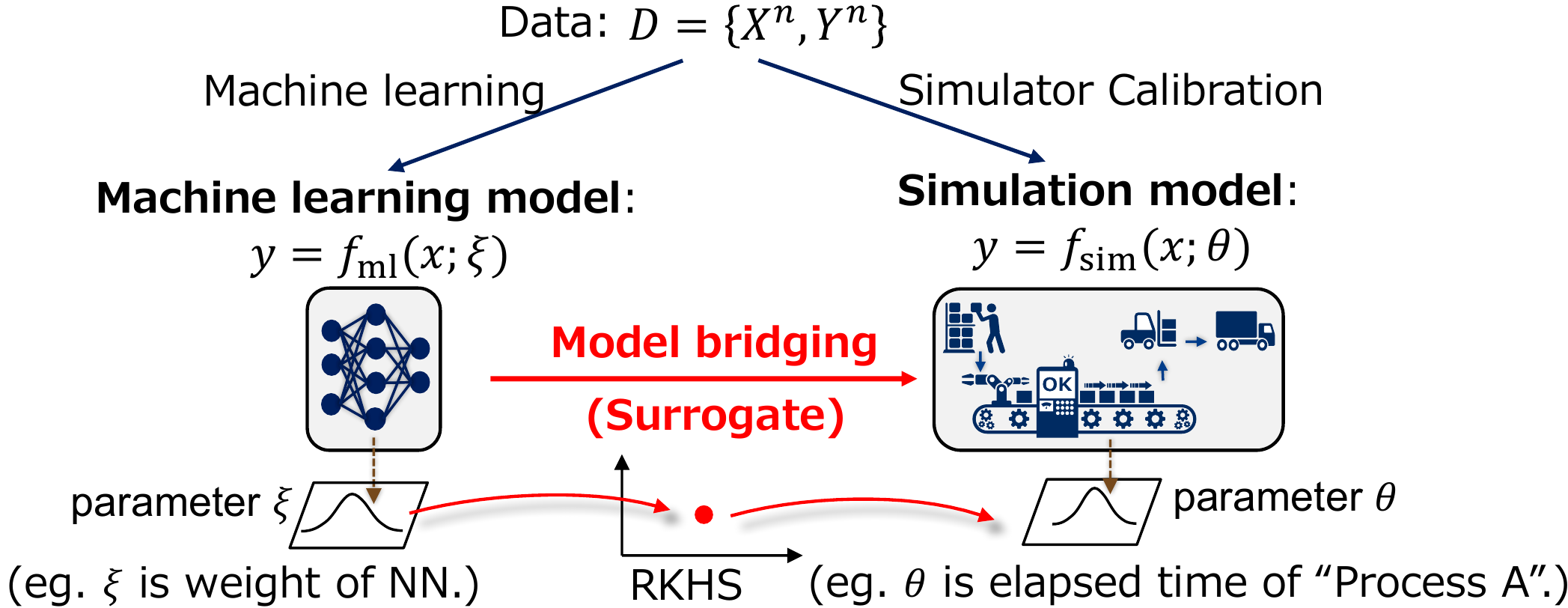}
\caption{Basic idea of the model-bridging framework.}
\vspace{-0.1cm}
\label{fig:overview}
\end{figure}

We propose a ``model-bridging (MB) '' framework to predict using a machine learning model as well as obtain interpretable simulation parameters simultaneously without expensive calculation of a simulation model.
The idea of this framework is to map the un-interpretable parameters of the machine learning model and the interpretable parameters of the simulation model (Fig.~\ref{fig:overview}).
The algorithm has to learn the relation of the posterior distribution estimated from each dataset between the machine learning model and the simulation model in advance; this framework can be considered as a meta-learning for each dataset.

Let us consider the example of production simulation for predicting the efficiency of manufacturing production, implementing a series of processes for production (example in Fig.~\ref{fig:simpleassembly_experiment}).
The production simulation aims to obtain a production efficiency and the reason for it simultaneously within a minute to improve the production efficiency.
We formulate this problem setting.
Assume that we obtain the dataset $\{X^n, Y^n\} = \{ X_1, ..., X_n, Y_1, ..., Y_n\} $, where input $X_i \in \mathbb{R}^{d_x}$ is the number of products to be manufactured in unit time and output $Y_i \in \mathbb{R}^{d_y}$ is the efficiency of production.
The simulation parameter $\theta \in \mathbb{R}^{d_{\theta}}$ is the elapsed time for each process, which undergoes a probabilistic behavior.
The parameter $\theta $ is interpretable and helpful in understanding the system and decision making.
Thus, we need to obtain the prediction $\hat{Y}_{n+1}$ for new data $X_{n+1}$ as well as obtain the interpretable simulation parameters $\theta$ representing the elapsed time of each process, which provides information regarding the occurrence of ``bottleneck processes.''
Here, the observed data and its generation process are considered to drift gradually, for example, the daily production of the factory due to the load of labors and machine environment factors such as temperature.
The detailed assumption is described in a later section.

Note that this study considered a different problem setting from the conventional methods with simplified surrogate models, such as LIME, SHAP, and rule-based model; the interpretable model of the proposed method, i.e., simulation model, is complex and computationally expensive.
There is no existing method for solving this new problem setting, where it is difficult to show the baseline for the evaluation.
Experimentally, we confirm that the estimation of model bridging is reasonable in comparison with simulator calibration as a baseline with a significantly fast process owing to no execution of the simulation.

The main contribution of this paper is to propose a novel framework for bridging machine learning and simulation, which has never been discussed before from the context of machine learning and to demonstrate its effectiveness in real-world applications.
The technical contribution is to expend the distribution-to-distribution regression on reproducing kernel Hilbert space (RKHS), as a suitable method for bridging function.
The rest of this paper is organized as follows. 
We briefly review a series of applications of kernel mean embedding as the building blocks for the proposed framework.
Subsequently, we propose the model-bridging framework.
Finally, we confirm the accuracy of the proposed method for three cases of simulation.

\section{Related Works}
\label{sec:RalatedWorks}

We briefly introduce simulator calibration and distribution regression based on kernel mean embedding as a building block of the proposed framework.

\subsection{Simulator Calibration}
\label{subsec:SimulatorCalibration}

``Simulator calibration''~\cite{Kisamori2018} is a method for estimating the simulation parameter as the posterior distribution to reproduce real data.
Simulator calibration is an example of data assimilation.
The simulation model is treated as a regression function $f_{\rm sim}(x; \theta)$ by combining a series of kernel mean embedding methods.
The conventional statistical methods of parameter estimation are not applicable to simulator calibration owing to the properties of the likelihood function: intractable or nondifferentiable.
When Gaussian noise is employed with regression function $f_{\rm sim}(x; \theta)$, the likelihood is expressed as
\begin{eqnarray}
\vspace{-0.1cm}
p(y|x, \theta) = \frac{1}{\sqrt{2\pi \sigma_0^2}} \exp \left\{ -\frac{1}{2\sigma_0^2} \left\| y - f_{\rm sim}(x; \theta) \right\| ^2 \right\}, \nonumber
\vspace{-0.1cm}
\end{eqnarray}
where $\sigma_0 > 0$ is a constant of observation noise.
This likelihood function is nondifferentiable owing to the simulation model $f_{\rm sim}(x; \theta)$.
The posterior mean to be obtained is formulated as $p(\theta | X^n, Y^n) = p(Y^n|X^n, \theta) \pi(\theta)/Z(X^n, Y^n)$, where $\pi(\theta)$ is the prior distribution and $Z(X^n, Y^n)$ is the regularization constant.
In this application, simulator calibration estimated the simulation parameter $\theta$ as a kernel mean of the posterior distribution by using kernel approximated Bayesian computation (kernel ABC)~\cite{Nakagome2013,Fukumizu2013}.
After obtaining the kernel mean of the posterior distribution, a posterior sample is obtained using kernel herding~\cite{Chen2010}.

\begin{figure}[t]
\centering
\vspace{-0.1cm}
\includegraphics[width=5.3cm, bb=0 0 366 123]{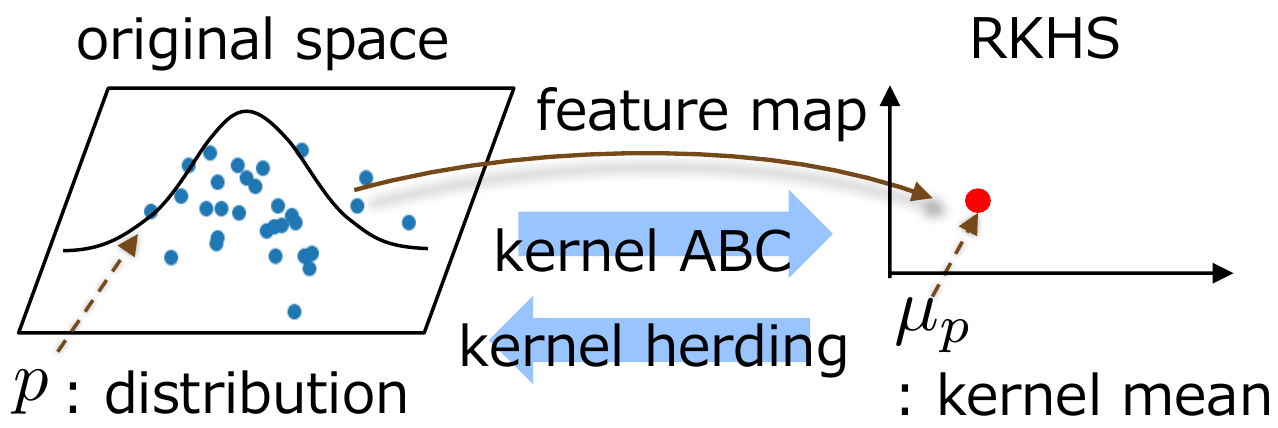}
\caption{Schematics of kernel mean embedding as a tool of model bridging.}
\vspace{-0.1cm}
\label{fig:kernel_mean_embedding}
\end{figure}

\subsection{Application of Kernel Mean Embedding}

As an application of kernel mean embedding~\cite{Muandet2016}, we briefly review the kernel ABC and kernel herding.
The kernel mean embedding is a framework for mapping distributions into a RKHS $\mathcal{H}$ as a feature space.
Kernel herding is a sampling method from the embedded distribution in RKHS that has the opposite operation of kernel mean embedding.
Figure~\ref{fig:kernel_mean_embedding} shows a schematic of the relation of kernel ABC and kernel herding.

{\bf Kernel ABC}: 
Kernel ABC~\cite{Nakagome2013,Fukumizu2013} is a method for computing the kernel mean of the posterior distribution from a sample of parameter $\theta$, generated by the prior distribution $\pi(\theta)$.
The assumption is that the explicit form of the likelihood function is unavailable, while the sample from the likelihood function is available.
The kernel ABC allows us to calculate the kernel mean of the posterior distribution as follows. First, the sample $\{\theta_1, ..., \theta_m\}$ is generated from prior distribution $\pi (\theta )$ and pseudo-data $\{\bar{Y}_1 ^n, ... , \bar{Y}_m^n \}$, as a sample from $p(y|x, \theta_j)$ for $j = 1, ..., m$.
Next, the empirical kernel mean of the posterior distribution 
\begin{eqnarray}
\hat{\mu}_{\theta | YX}=\sum_{j=1}^m w_j k_{\theta}(\cdot, \theta_j) \label{eq:mu_sim}
\end{eqnarray}
is calculated, where $k_{\theta}$ is a kernel of $\theta$. Weight $w_j$ is calculated by 
\begin{eqnarray}
(w_1, ..., w_m)^T &=& (G_y+m\delta I)^{-1} {\bf{k}}_y(Y^n) \ \in \mathbb{R}^m \nonumber \\
G_y &=& \{ k_y(\bar{Y}^n_j, \bar{Y}^n_{j\prime}) \} _{j, j\prime =1} ^m \ \in \mathbb{R}^{m\times m} \label{eq:w} \\
{\bf{k}}_y (Y^n) &=& (k_y(\bar{Y}^n_1, Y^n), ..., k_y(\bar{Y}^n_m, Y^n)) \ \in \mathbb{R} ^m. \nonumber
\end{eqnarray}
The $\delta \geq 0$ is a regularization constant, $I$ is an identity matrix, and $k_y$ is a kernel of $y$.
The kernel $k_y(\bar{Y}^n_j, Y^n)$ indicates the ``similarity'' between pseudo-data $\bar{Y}_j^n$ and real data $Y^n$.
The calculations of the kernel mean corresponds to the estimation of the posterior distribution as an element in $\mathcal{H}$.

{\bf Kernel Herding}: 
Kernel herding~\cite{Chen2010} is a method used for sampling data from the kernel mean representation of a distribution, which is an element of the RKHS. Kernel herding can be considered as an opposite operation to that of kernel ABC.
Kernel herding greedily obtains samples of $\theta$ by updating Eqs.(1) and (2) as given in Chen et al.~\cite{Chen2010}.

\subsection{Distribution Regression}

Distribution regression is a regression for $d_x$-dimensional ``distributions'' represented by samples.
In contrast, normal regression is regression for $d_x$-dimensional ``point.''
There are several studies of distribution regression, including distribution-to-distribution regression~\cite{Oliva2013} and distribution-to-point regression~\cite{Szabo2016,Law2017}.
Oliva et al.~\cite{Oliva2013} employed the idea of approximating a density function by kernel density estimation, rather than using RKHS.
Szab\'{o} et al.~\cite{Szabo2016} proposed the distribution-to-point with the kernel ridge regression method on RKHS; however, no methods are available for distribution-to-distribution regression on RKHS.

\begin{figure}[t]
\centering
\vspace{-0.1cm}
\includegraphics[width=8.cm, bb=0 0 469 473]{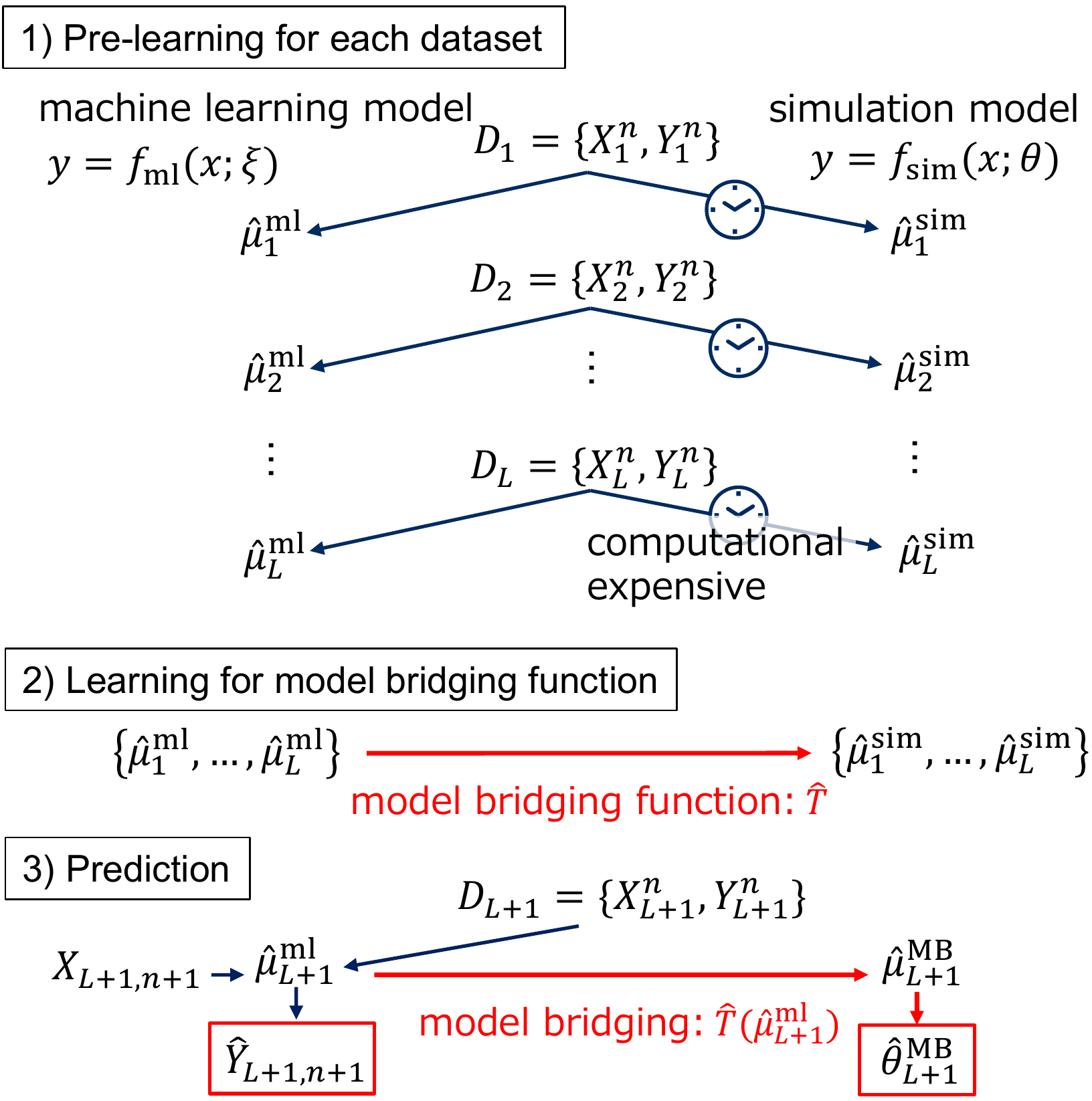}
\caption{Overview of the algorithm of the model-bridging framework.}
\vspace{-0.1cm}
\label{fig:framework}
\end{figure}

\begin{table*}[t]
\centering
\vspace{-0.1cm}
\begin{tabular}{l}
\hline
{\bf Algorithm 1}: Model bridging \\
\hline
\small    {\it1) Pre-learning for each dataset} : \\
\small    {\bf Input}: Dataset $\{X^n_l, Y^n_l \} _{l=1}^L$, \\
\small    \hspace{1cm} machine learning model $f_{\rm ml}(x, \xi)$\\ 
\small    \hspace{1cm} and simulation model $f_{\rm sim}(x, \theta)$ \\
\small    {\bf Output}: $\{ \hat{\mu}^{\rm ml}_1, ..., \hat{\mu}^{\rm ml}_L\}$ and $\{ \hat{\mu}^{\rm sim}_1, ..., \hat{\mu}^{\rm sim}_L\}$ \\
\small    {\bf for} $l=1$ to $L$ {\bf do}\\
\small        \hspace{1cm} Estimation for $\hat{\mu}^{\rm ml}_l$ by Eq. (\ref{eq:mu_y_given_x})\\
\small        \hspace{1cm} Estimation for $\hat{\mu}^{\rm sim}_l$ by Eq. (\ref{eq:mu_sim})\\
\small    {\bf end for}\\
\small    {\it 2) Learning for model-bridging function $\hat{T}$: } \\
\small    {\bf Input}: $\{ \hat{\mu}^{\rm ml}_1, ..., \hat{\mu}^{\rm ml}_L\}$ and $\{ \hat{\mu}^{\rm sim}_1, ..., \hat{\mu}^{\rm sim}_L\}$ \\
\small    {\bf Output}: Model-bridging function $\hat{T}$ \\
\small    Learning for $\hat{T}$ by Eq. (\ref{eq:T}) \\
\small    {\it 3) Prediction: } \\
\small    {\bf Input}: Dataset $\{ X^n_{L+1}, Y^n_{L+1} \}$ and $X_{L+1, n+1}$\\
\small    {\bf Output}: $\hat{Y}_{L+1, n+1}$ and $\hat{\theta}_{L+1}$\\
\small    Estimation for $\xi_{L+1}$ and $\hat{\mu}^{\rm ml}_{L+1}$ \\
\small    Prediction for $\hat{Y}_{L+1, n+1}$ by $f_{\rm ml}(x; \xi_{L+1})$ \\
\small    Estimation for $\hat{\mu}^{\rm MB}_{L+1}=\hat{T}(\hat{\mu}^{\rm ml}_{L+1})$ by Eq. (\ref{eq:mu_sim_Lp1})\\
\small    Sampling for $\hat{\theta}^{\rm MB}_{L+1}$ by Eq. (\ref{eq:update_kernel_herding})\\
\hline
\end{tabular}
\vspace{-0.1cm}
\end{table*}

\section{Proposed Framework: Model Bridging}

We propose a novel framework to bridge the un-interpretable machine learning model and the interpretable simulation model.
In this study, we assume a machine learning model, such as a Bayesian neural network (BNN)~\cite{Neal1996}, as a parametric model and a Gaussian process as a non-parametric model.
This proposed framework is applicable to any model.
In this section, first, we confirm the problem setting and framework of model bridging.
Second, we propose the algorithm of distribution-to-distribution regression, which is suitable for the proposed framework.
Thereafter, we propose the formulation of the input of distribution-to-distribution regression for the parametric model, assuming BNN, and the non-parametric model, assuming the Gaussian process.
Figure~\ref{fig:framework} and Alg. 1 shows an overview of the framework.

\subsection{Problem Setting, Assumption, and Usage of Model Bridging}
\label{subsec:ProblemSetting}

We define the problem setting of the model-bridging framework.
Let $L$ be dataset $\{X_{l}^n, Y_{l}^n\}_{l=1}^L $ $ (X^n_l \in \mathbb{R}^{n \times d_x}, Y^n_l \in \mathbb{R}^{n \times d_y})$, given in the pre-learning phase.
For simplicity of explanation, we use the unique number of the data $n$ and sample size $m$ for all datasets.
However, it can be different numbers generally, such as $n_l$ and $m_l$.
The purpose is to predict $\hat{Y}_{L+1, n+1}$ and simultaneously obtain interpretable simulation parameter $\hat{\theta}^{\rm MB}_{L+1}$ to reproduce $Y_{L+1, n+1} = f_{\rm sim}(X_{L+1, n+1}; \hat{\theta}^{\rm MB}_{L+1})$ without the expensive calculation of simulation model $f_{\rm sim}(x; \theta)$, when we obtain new dataset $\{ X^n_{L+1}, Y^n_{L+1} \}$.
The assumptions of the problem setting are as follows.
These assumptions are prevalent for many applications of a simulation.
\begin{itemize}
\item The existing simulation model $f_{\rm sim} (x; \theta)$ with interpretable simulation parameter $\theta \in \mathbb{R}^{d_{\theta}}$ and a machine learning model $f_{\rm ml}(x; \xi)$ that is sufficiently accurate to predict a typical regression problem while having un-interpretable parameter $\xi \in \mathbb{R}^{d_{\xi}}$.
\item The cost of simulator calibration is much higher than that of learning for the machine learning model. 
For instance, it takes more than one hour for simulator calibration of one dataset $\{ X^n_l, Y^n_l\}$, while learning of BNN takes less than a minute.
\item Dataset $\{X_l^n, Y_l^n \}$ has dependency of parameter $\theta_l$ for each $l=1,..., L$. 
Let us assume the following situation. $\{ X^n_l, Y^n_l\}$ is obtained in one day with the same conditions, described as parameter $\theta_l$, while the conditions are changed for the following day, described as $\theta_{l+1}$.
\item The time for offline calculation of simulator calibration is sufficient, while the time for prediction is restricted.
\end{itemize}

Once we obtain the model-bridging function as a mapping from the machine learning model to the simulation model, we can obtain an accurate prediction for $\hat{Y}_{L+1, n+1}$ by both the machine learning model and interpretable $\hat{\theta}^{\rm MB}_{L+1}$ by the simulation model for new dataset $\{ X^n_{L+1}, Y^n_{L+1} \}$ without an expensive calculation from the simulation model.

\subsection{Distribution-to-Distribution Regression}

We present the regression algorithm between the conditional kernel mean of the machine learning model $\mu^{\rm ml} \in \mathcal{H}$ and that of the simulation model $\mu^{\rm sim} \in \mathcal{H}$, as a model-bridging function $\mu^{\rm sim} = T(\mu^{\rm ml})$.
We develop the algorithm based on kernel ridge regression, which is suitable for kernel mean input and output on RKHS.
This is the extension of the distribution-to-point regression method proposed by Szab\'{o} et al.~\cite{Szabo2016} for the distribution output.

\subsubsection{Kernel Ridge Regression for Kernel Mean}

The formulation to be solved as an analogy of normal kernel ridge regression is as follows:
\begin{eqnarray}
\hat{T} = {\rm arg} \max _{T \in \mathcal{F}} \frac{1}{L} \sum_{l=1}^L \| \hat{\mu}^{\rm sim}_{l} - T(\hat{\mu}^{\rm ml}_{l}) \|^2_{\mathcal{F}} + \lambda \| T \|^2 _{\mathcal{F}}, \label{eq:T}
\end{eqnarray}
where $\lambda \geq 0$ is a regularization constant.
$\mathcal{F}$ is a function space of kernel mean embeddings following Christmann et al. \cite{Christmann2010} and $\| \cdot \|_{\mathcal{F}} $ is its norm.
The difference from ordinary kernel ridge regression is that the inputs and outputs are kernel means.
Therefore, we define kernel $\kappa \in \mathcal{F}$, as a function of kernel mean $\mu \ \in \mathcal{H}$.
We employ a Gaussian-like kernel as 
\begin{eqnarray}
\kappa (\mu, \mu^{\prime}) = \exp \left\{ -\frac{1}{2\sigma_{\mu}^2} \left\| \mu - \mu^{\prime} \right\|_{\mathcal{H}} ^2 \right\} \ \in \mathcal{F}, \label{eq:kappa}
\end{eqnarray}
where constant $\sigma_{\mu} > 0$ is the width of kernel $\kappa$ and $\| \cdot \| _{\mathcal{H}}$ is RKHS norm.
The kernel $\kappa$ is also a positive definite kernel~\cite{Christmann2010}.

Following the representor theorem of kernel ridge regression~\cite{Kung2014}, the estimated model-bridging function $\hat{T}$ for new $\hat{\mu}^{\rm ml}_{L+1}$ is described as 
\begin{eqnarray}
\hat{\mu}^{\rm MB}_{L+1} = \hat{T}(\hat{\mu}^{\rm ml}_{L+1}) = \sum_{l=1}^{L} v_l \hat{\mu}^{\rm sim}_{l} \ \ \in \mathcal{F}, \label{eq:mu_sim_Lp1}
\end{eqnarray}
where ${\bf v} = (v_1, ..., v_L )^T = (G_{\mu} + \lambda LI )^{-1} {\bf k}_{\mu}(\hat{\mu}^{\rm ml}_{L+1}) \ \  \in \mathbb{R} ^L$.
Gram matrix $G_{\mu}$ and the vector ${\bf k}_{\mu}(\hat{\mu}^{\rm ml}_{L+1})$ are described as follows:
\begin{eqnarray}
G_{\mu} &=& \left\{ \kappa (\hat{\mu}^{\rm ml}_{l}, \hat{\mu}^{\rm ml}_{l^{\prime}} ) \right\} _{l, l^{\prime} = 1} ^{L} \ \ \in \mathbb{R}^{L \times L} \nonumber \\
{\bf k}_{\mu}(\hat{\mu}^{\rm ml}_{L+1}) &=& \left( \kappa (\hat{\mu}^{\rm ml}_{1}, \hat{\mu}^{\rm ml}_{L+1}), ..., \kappa (\hat{\mu}^{\rm ml}_{L}, \hat{\mu}^{\rm ml}_{L+1}) \right)^T \ \in \mathbb{R}^L. \nonumber
\end{eqnarray}

\subsubsection{Kernel Herding from Kernel Mean $\hat{\mu}^{\rm MB}_l$}

After obtaining the kernel mean of $\hat{\mu}^{\rm MB}_{L+1}$, kernel herding can be applied to sample $\hat{\theta}^{\rm MB}_{L+1} = \{ \hat{\theta}_{L+1, 1},..., \hat{\theta}_{L+1, m} \} $, where $\hat{\theta}_{L+1, j} \in \mathbb{R}^{d_{\theta}}$.
The explicit form of the update equation for sample $j=1,...,m$ iteration of kernel herding with kernel mean $\hat{\mu}^{\rm MB}_{L+1}$ is as follows:
\vspace{-0.1cm}
\begin{eqnarray}
\hat{\theta}_{L+1, j} = {\rm arg}\max_{\theta} \sum_{l=1}^L \sum_{j^{\prime}=1}^m v_l w_{l,j^{\prime}} k_{\theta}(\theta, \theta_{l, j^{\prime}}) - \frac{1}{j}\sum_{j^{\prime}=1}^{j-1} k_{\theta}(\theta, \theta_{j^{\prime}}) \ \in \mathbb{R}^{d_{\theta}}, \label{eq:update_kernel_herding}
\end{eqnarray}
for $j=2, ..., m$.
For initial state $j=1$, the update equation constitutes only the first term of Eq. (\ref{eq:update_kernel_herding}).
The weight of $w_{l,j}$ is calculated by kernel ABC for dataset $\{ X^n_l, Y^n_l \}$ in Eq.(\ref{eq:w}).

\subsection{Input of Distribution-to-Distribution Regression}

We present the explicit formulation for calculating the kernel means of the machine learning model $\hat{\mu}^{\rm ml}_{l}$, as an input of the distribution-to-distribution regression.
First, we present the formulation of BNN, as BNN is a useful model for many applications as a parametric Bayesian model.
Second, we present the formulation for Gaussian process regression as a non-parametric Bayesian model. 
We consider the Gaussian process regression as a non-parametric alternative to BNN.
The equivalence between the Gaussian process and BNN with one hidden layer with infinite nodes is well known~\cite{Neal1996}.
Furthermore, a recent study reveals the kernel formulation that is equivalent to multi-layered BNN, as an extension of the Gaussian process~\cite{Lee2018}.
We directly obtain empirical kernel mean without calculation of kernel from parameters in the parametric model for using Gaussian process regression.

\subsubsection{Parametric Model: Bayesian Neural Network}

We assume the BNN model $f_{\rm ml} (x; \xi)$ with a few hidden layers, where $\xi$ is a parameter, such as weights for each node and bias terms of each layer.
We can obtain the posterior distribution of $\xi_{l} $ for $l=1,...,L$ by the Markov Chain Monte Carlo method or variational approximation.
Then, the empirical kernel mean of the posterior distribution is represented as $\hat{\mu}^{\rm ml}_l = \sum_{j=1}^m k_{\xi} (\cdot, \xi_{l, j}) \ \in \mathcal{H}$ for $l=1, ..., L$ dataset, where $k_{\xi}$ is kernel of $\xi$.

We employ Gaussian-like kernel $\kappa$ as an function of $\hat{\mu}^{\rm ml}_{l} \in \mathcal{H}$ as
\begin{eqnarray}
\kappa (\hat{\mu}^{\rm ml}_{l}, \hat{\mu}^{\rm ml}_{l^{\prime}}) = \exp \left\{ -\frac{1}{2\sigma_{\mu}^2} \left\| \hat{\mu}^{\rm ml}_{l} - \hat{\mu}^{\rm ml}_{l^{\prime}} \right\|_{\mathcal{H}} ^2 \right\} \ \in \mathcal{F} \nonumber \\
= \exp \left\{ -\frac{1}{\sigma^2_{\mu}} \left( 1-\sum_{j=1}^m \sum_{j^{\prime}=1}^{m^{\prime}} k_{\xi}(\xi_{l, j}, \xi_{l^{\prime}, j^{\prime}}) \right) \right\}. \nonumber
\end{eqnarray}
The relation $\langle \hat{\mu}^{\rm ml}_{l}, \hat{\mu}^{\rm ml}_{l^{\prime}} \rangle =\sum_{j=1}^m \sum_{j^{\prime}=1}^{m^{\prime}} k_{\xi}(\xi_{l, j}, \xi_{l^{\prime}, j^{\prime}}) $  is used, where $\langle \cdot , \cdot \rangle$ represents the inner product.

\subsubsection{Non-Parametric Model: Gaussian Process Regression}

We use the Gaussian process regression as a non-parametric model. 
In this case, we can directly obtain empirical kernel mean without the calculation of kernel from parameters, such as $\xi$ in the parametric model.
We present that the prediction with Gaussian process regression can be considered as a conditional kernel mean.
As a result of Gaussian process regression, we can express the mean of the predictive distribution in general for $l$-th dataset $\{X^n, Y^n \}$ as 
\begin{eqnarray}
y = \hat{\mu}_{Y|X, l} (x) = \sum_{i=1}^n u_{l, i}(x) k_y(\cdot , Y_{l, i}), \label{eq:mu_y_given_x}
\end{eqnarray}
where $u_{l,i}(x) = \{ (G_x + n \lambda^{\prime} I)^{-1} {\bf k}_x(x)\}_{i}$. 
The $G_x$ is the Gramm matrix, $\lambda ^{\prime} \geq 0$ is regularization constant, and $k_x$ is kernel of $x$.
This formulation is clear if we remember the equivalence between Gaussian process regression and kernel ridge regression~\cite{Kanagawa2018}.
As a predictor of $\hat{Y}_{l, n+1}$ for new $X_{l, n+1}$, we can calculate $\hat{Y}_{l, n+1}=\hat{\mu}_{Y|X, l}(X_{l,n+1})$.
Note that this $\hat{\mu}_{Y|X, l}$ is interpreted as the kernel mean of the $Y^n_l$ conditioned by $X^n_l$.
Thus, we can use $\hat{\mu}_{Y|X,l}$ as the input of the distribution-to-distribution regression, represented as $\hat{\mu}^{\rm ml}_l$.
We employ Gaussian-like kernel $\kappa$ as an function of $\hat{\mu}^{\rm ml}_l$ as
\begin{eqnarray}
\kappa (\hat{\mu}^{\rm ml}_{l}, \hat{\mu}^{\rm ml}_{l^{\prime}})  = \exp \left\{ -\frac{1}{\sigma^2_{\mu}} \left( 1-\sum_{i=1}^n \sum_{i^{\prime}=1}^{n^{\prime}} u_{l,i}u_{l^{\prime}, i^{\prime}} k_{y}(Y_{l, i}, Y_{l^{\prime}, i^{\prime}}) \right) \right\}. \nonumber
\end{eqnarray}
There is a difference between the proposed non-parametric method and parametric method. 
In the parametric method, the distribution of parameter $\xi_l$ is the input of the distribution-to-distribution regression, while in the non-parametric method, the distribution of data $Y^n_l$ conditioned by $X^n_l$ is the input.

\section{Experiment}
\label{sec:Experiment}

This framework is widely applicable to various domains for industries that include multi-agent simulation, traffic simulation, and simulation of dynamics of physical systems, such as thermomechanics, structural mechanics, and electromagnetic mechanics.
We present the applications of the model-bridging framework for three simulations: 1) a simple production simulation to show and explain the framework effectiveness in detail, 2) a realistic production simulation to show the capability for realistic application, and 3) a simulation of fluid dynamics to show of the applicability to a wide variety of simulation fields.
The detailed information on the three experiments is provided in the supplemental material owing to page limitations.

Through the three experiments, we confirm that model bridging enables us to predict $\hat{Y}_{L+1, n+1}$ for new $X_{L+1, n+1}$, using a machine learning model, and obtain the interpretable simulation parameter $\hat{\theta}^{\rm MB}_{L+1}$ without an expensive calculation of the simulation when we obtain the $L+1$-th dataset $\{ X^n_{L+1}, Y^n_{L+1}\}$.
We also investigate the accuracy of the estimation of the parameter $\hat{\theta}^{\rm MB}_{L+1}$ and compared the execution time with simulator calibration as a baseline.
Note that these experiments cannot be compared with other state-of-the-art surrogate approaches, such as LIME, SHAP, and other methods with model compression, because no other methods exist that simultaneously obtain the prediction result and interpretable simulation parameters.

\subsection{Common Setting of Experiments}

In practice, the effective hyperparameter for model-bridging function to be tuned is the regularization constant $\lambda$ for distribution-to-distribution regression.
The hyperparameter $\lambda $ can stabilize the calculation of the inverse Gram matrix.
This hyperparameter should be determined by cross-validation.
Further, as a common hyperparameter of the kernel method, the width of the kernel must be selected to measure the similarity between the data.
We employed a Gaussian kernel for $k_y$, $k_x$, $k_{\theta}$, and $k_{\xi}$ for all experiments.
The typical setting of the width of the kernel, in practice, is the median of Euclid distance of the input data of a kernel.
In all the experiments performed in this study, we apply this setting and confirmed that all kernels perform adequately. 
We used a PC equipped with a 3.4-GHz Intel core i7 quad-core processor and 16GB memory.
The main computational cost is for a simulation in the pre-learning phase of this framework.

\subsection{Experiment with Simple Production Simulator}
\label{subsec:SimpleExperiment}

\subsubsection{Setting}

\begin{figure}[t]
  \centering
  \vspace{-0.1cm}
  \includegraphics[width=5.8cm, bb=0 0 361 125]{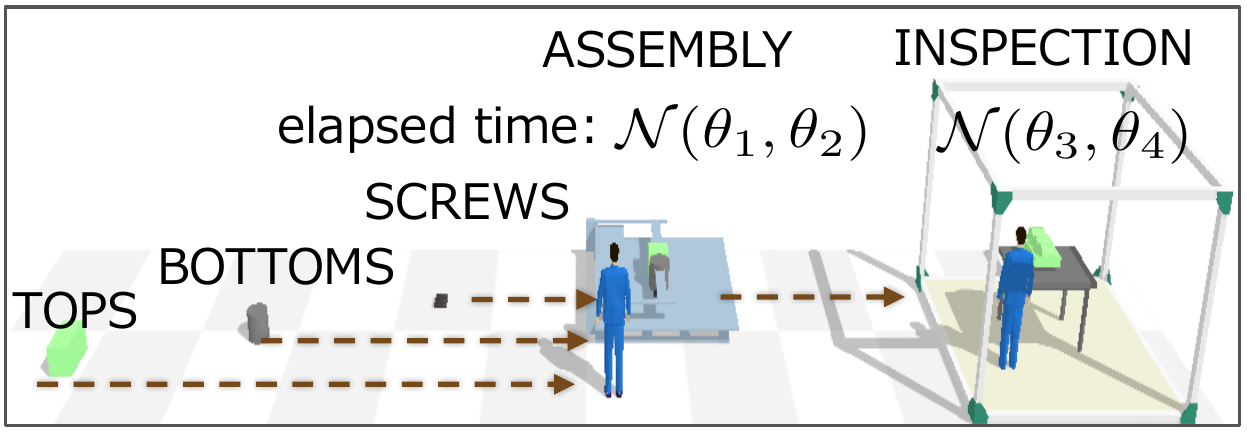}
  \caption{Production simulation model for the experiment.}
  \label{fig:simpleassembly_experiment}
  \vspace{-0.1cm}
\end{figure}

Production simulators are widely used simulation software for discrete and interconnection systems to model various processes, such as production, logistics, transportation, and office works.
We used a {\em WITNESS}, a popular software package of production simulation\footnote[1]{\url{https://www.lanner.com/en-us/technology/witness-simulation-software.html}}.
We examined the regression problem using a simulation model that has a simple four-dimensional simulation parameter $\theta \in \mathbb{R}^4$ (Fig.~\ref{fig:simpleassembly_experiment}).
We defined the simulation input $x = X_i \in \mathbb{R}$ as the number of products to be manufactured, output $Y_i = f_{\rm sim}(X_i , \theta) \in \mathbb{R}$ as the total time to manufacture all the $X_i$-th products, and parameter $\theta$ as the time required for each procedure on the production line.
Moreover, the time required for ``ASSEMBLY'' is $\mathcal{N}(\theta_1, \theta_2)$, and that for ``INSPECTION'' is $\mathcal{N}(\theta_3, \theta_4)$, where $\mathcal{N}(\mu _{\rm ND}, \sigma_{\rm ND})$ is the normal distribution with mean $\mu_{\rm ND}$ and standard deviation $\sigma_{\rm ND}$.
We assumed that the elapsed time of each process would increase considerably, owing to an increasing load, if the number of products to be manufactured also increases.
To create this situation artificially, we set different true parameters between the observed data region $\theta^{(0)}$ and the predictive region $\theta^{(1)}$.
We set $\theta^{(0)} = (2, 0.5, 5, 1 )^T$ if $x \leq 110$ and $\theta^{(1)} = (3.5, 0.5, 7, 1)^T$ if $x > 110$.
The shift in parameters $\theta_1$ and $\theta_3$ between $\theta^{(0)}$ and $\theta^{(1)}$ is the sigmoid function.
For each $l$-th dataset, the observed data of size $n=50$ and sample size $m=100$ are generated by $q_l(x) = \mathcal{N}(\chi _l, 5)$, where $\chi_l$ is uniform distribution in $[70, 130]$ for $l=1, ..., L$.
The number of training datasets, $L$, is 100.
We defined the prior distribution as the uniform distribution over $[0,5] \times [0,2] \times [0,10] \times [0,2]$.
We used a BNN having two fully connected hidden layers with three nodes and bias nodes for each layer, as a machine learning model. The activation function is ReLU.
The regularization constant is $\lambda = 1.0 \times 10^{-6}$ for this experiment.

\subsubsection{Result}

\begin{figure}[t]
\centering
\vspace{-0.1cm}
\includegraphics[width=9.8cm, bb=0 0 472 210]{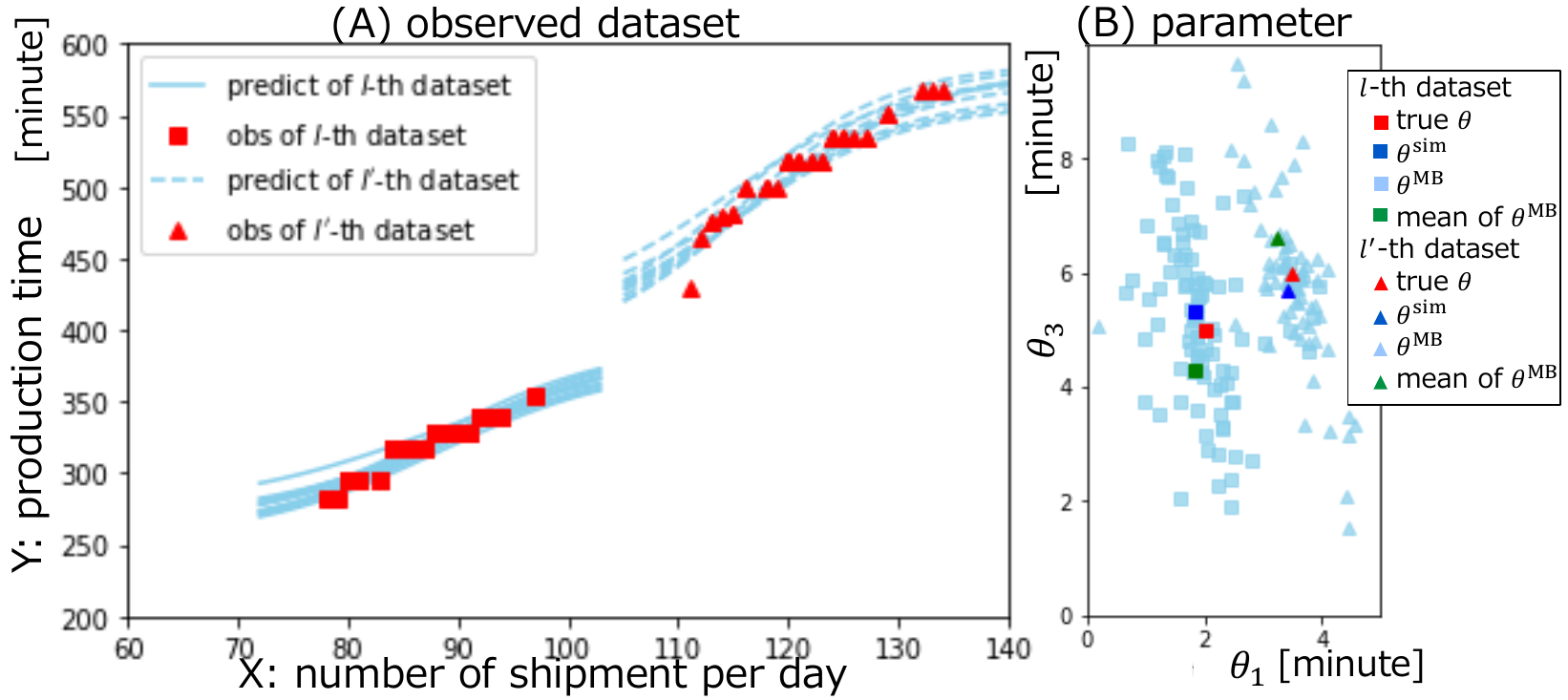}
\caption{As representatives for all dataset, two test datasets are shown: $l$-th dataset as square markers and $l^{\prime}$-th dataset as triangle markers. (A) Observed data and fitted result by BNN. (B) Estimated distribution of simulation parameters by model bridging.}
\vspace{-0.1cm}
\label{fig:simpleassembly_experiment_result}
\end{figure}

\begin{figure}[t]
\centering
\vspace{-0.1cm}
\includegraphics[width=5.6cm, bb=0 0 303 206]{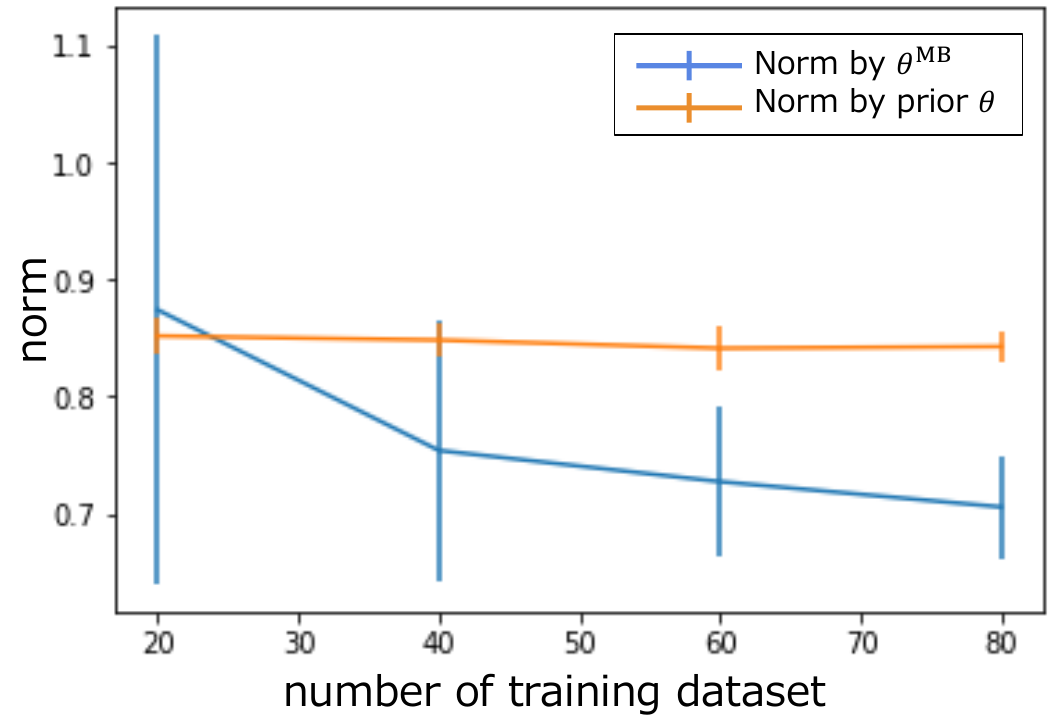}
\caption{Estimation of norm for $\hat{\mu}$ by the model-bridging function for number of training datasets. The blue line represents the estimated result of the norm, and the orange line represents the norm with prior information only.}
\vspace{-0.1cm}
\label{fig:norm_mu}
\end{figure}

The execution time of model bridging is 9.6 [s] in the presented computational environment, while simulator calibration requires about 3.1 [h] for $L+1$-th dataset.
Simulator calibration requires $m \times n$ execution of simulation and each simulation takes 2 [s] in this case.
As representatives of all test datasets, the results of the model-bridging framework for two different datasets ($l$-th and $l^{\prime}$-th dataset), which are randomly selected, are shown in Fig.~\ref{fig:simpleassembly_experiment_result}.
Figure~\ref{fig:simpleassembly_experiment_result}~(A) shows the observed data for $l$-th dataset as red squares and the $l^{\prime}$-th dataset as red triangles.
The solid line and dashed line are the fitted results by BNN with variational approximation.
Figure~\ref{fig:simpleassembly_experiment_result}~(B) shows the estimated posterior distributions of simulation parameters by model bridging $\hat{\theta}^{\rm MB}_l$ each dataset.
The red markers show the true parameter, green markers show the estimated result of simulator calibration, and blue markers show the mean of the distribution.
Each square denotes the $l$-th dataset and each triangle denotes the $l^{\prime}$-th dataset.
We can see a reasonably accurate estimation of $\theta_l ^{\rm MB}$ and $\theta_{l^{\prime}}^{\rm MB}$ model-bridging framework in comparison with simulator calibration for the two different datasets with two different true parameters $\theta$.

Note that from the perspective of interpretability, we can clearly see the practical effectiveness of simultaneously obtaining the prediction result with interpretable parameters, such as ``elapsed time of a process.''
From these interpretable parameters, we can understand that the production efficiency is decreased ($l^{\prime}$-th dataset in Fig.\ref{fig:simpleassembly_experiment_result}) mainly because of the increased elapsed time of ``INSPECTION'' ($= \theta_1$).

We also investigated $\| \hat{\mu}^{\rm MB}_{L+1} - \hat{\mu}^{\rm sim}_{L+1} \| ^2 _{\mathcal{H}}$ to confirm the convergence of the proposed distribution-to-distribution regression in the model-bridging framework.
The detailed formulation for the numerical calculation is presented in the supplementary material.
Figure~\ref{fig:norm_mu} shows the mean and standard deviation of one-leave-out cross-validation of the test dataset. 
The horizontal axis shows the number of training datasets.
We can see the convergence for bias that originates from simulator calibration.

\subsection{Experiment with Realistic Production Simulator}

\subsubsection{Setting}

\begin{table*}[t]
\centering
\caption{Summary of true and estimated parameters in the experiment for realistic production simulation. $T_{\rm BF}$ represents the mean time between failures, and $T_R$ represents the mode time of repair for each process. The estimated parameters are mean and standard deviation (in parentheses) of posterior mean for one-leave-out cross-validation.}
\vspace{-0.1cm}
\begin{tabular}{c|cc|cc|cc|cc|cc|cc}
Process  & \multicolumn{2}{c|}{Saw} & \multicolumn{2}{c|}{Coat}  & \multicolumn{2}{c|}{Inspection} & \multicolumn{2}{c|}{Harden} & \multicolumn{2}{c|}{Grind} & \multicolumn{2}{c}{Clean} \\ \hline
& $T_{\rm BF}$ & $T_{\rm R}$ & $T_{\rm BF}$ & $T_{\rm R}$ & $T_{\rm BF}$ & $T_{\rm R}$ & $T_{\rm BF}$ & $T_{\rm R}$ & $T_{\rm BF}$ & $T_{\rm R}$ & $T_{\rm BF}$ & $T_{\rm R}$ \\
Param. & $\theta_1$ & $\theta_2$ & $\theta_3$ & $\theta_4$ & $\theta_5$ & $\theta_6$ & $\theta_7$ & $\theta_8$ & $\theta_9$ & $\theta_{10}$ & $\theta_{11}$ & $\theta_{12}$ \\ \hline
\small $\theta^{(0)} $ & \small 100 & 25 & \small 150 & \small 5 & \small {\bf 100} & \small 20 & \small 150 & \small 5 & \small 75 & \small 15 & \small 120 & \small 20 \\
\small $\theta^{(1)} $ & 100 & \small 25 & \small 150 & \small 5 & \small {\bf 80}   & \small 20 & \small 150 & \small 5 & \small 75 & \small 15 & \small 120 & \small 20 \\ \hline
\small $\hat{\theta}^{\rm sim}$ & \small 100.5 & \small 25.1 & \small 153.2 & \small 5.0 & \small {\bf 104.2} & \small 17.3 & \small 146.5 & \small 5.1 & \small 73.7 & \small 14.6 & \small 96.8 & \small 20.2 \\ [-0.5pt]
$(x\leq 30)$ & \small (9.5) & \small (2.2) & \small (14.2) & \small (0.4) & \small (10.1) & \small (1.9) & \small (14.2) & \small (0.3) & \small (6.7) & \small (1.2) & \small (15.7) & \small (2.3)\\ [1pt]
\small $\hat{\theta}^{\rm sim}$ & \small 100.5 & \small 24.7 & \small 153.6 & \small 5.0 & \small {\bf 84.3} & \small 22.8 & \small 148.2 & \small 5.0 & \small 73.8 & \small 15.2 & \small 115.4 & \small 20.4 \\ [-0.5pt]
$(x > 30)$  & \small (8.3) & \small (1.4) & \small (10.6) & \small (0.3) & \small (7.4) & \small (0.9) & \small (8.9) & \small (0.3) & \small (4.6) & \small (1.0) & \small (9.8) & \small (1.8) \\ 
\hline
\small $\hat{\theta}^{\rm MB}$ & \small 102.5 & \small 21.2 & \small 165.1 & \small 6.0 & \small {\bf 98.6} & \small 17.0 & \small 179.0 & \small 5.5 & \small 74.0 & \small 17.8 & \small 94.6 & \small 12.7 \\ [-0.5pt]
$(x\leq 30)$ & \small (12.7) & \small (1.8) & \small (12.6) & \small (0.4) & \small (9.9) & \small (1.8) & \small (12.1) & \small (0.3) & \small (6.8) & \small (1.6) & \small (14.5) & \small (2.3) \\
\small $\hat{\theta}^{\rm MB}$ & \small 104.1 & \small 21.1 & \small 165.3 & \small 6.1 & \small {\bf 86.7} & \small 19.1 & \small 180.8 & \small 5.4 & \small 74.0 & \small 18.1 & \small 99.3 & \small 13.2 \\ [-0.5pt]
$(x > 30)$ & \small (10.1) & \small (2.3) & \small (14.2) & \small (0.5) & \small (11.8) & \small (2.1) & \small (8.1) & \small (0.5) & \small (4.5) & \small (1.2) & \small (21.2) & \small (1.6) \\
\end{tabular}
\label{table:ACME_parameters}
\vspace{-0.1cm}
\end{table*}

We used a model to reproduce a real metal-processing factory that manufactures valves from metal pipes, with six primary processes: ``saw,'' ``coat,'' ``inspection,'' ``harden,'' ``grind,'' and ``clean,'' in the order shown in the supplementary material.
Each process is composed of complex procedures, such as the preparation rule, waiting, and machine repair during trouble.
The purpose of this production simulation is also to predict the total production time $Y_i \in \mathbb{R}$ when the number of units $X_i \in \mathbb{R}^3$ for three types of products to be manufactured is set.
Each of the six processes contains two parameters of machine downtime owing to failure: mean time between failures ($T_{\rm BF}$) and mode time required for repair ($T_{\rm R}$).
We defined these parameters as twelve-dimensional parameter $\theta \in \mathbb{R}^{12}$ (see Table~\ref{table:ACME_parameters}).
The distribution of the mean time between failures is represented as a negative exponential distribution.
The distribution of the time required for repair is represented as an Erlang distribution with the mode time and shape parameter set at three.

Similar to the simple experiment discussed in the previous section, we set the true parameter as $\theta^{(0)}$ if $x \leq 30$ and $\theta^{(1)}$ if $x > 30$.
The summary of the true parameter is shown in Table~\ref{table:ACME_parameters}.
The shift in parameter $\theta_5$ between $\theta^{(0)}$ and $\theta^{(1)}$ is a sigmoid function.
The observed data of size is $n=30$ by $\mathcal{N}(\chi _l, 3)$ where $\chi _l$ is generated by uniform distribution in $[20, 40]$.
The number of parameter samples is $m=50$ and the number of datasets is $L=40$.
We defined the prior distribution as the uniform distribution over $[60,140] \times[15,35] \times [100,200] \times [3,10] \times [60,140] \times [15,35] \times [100,200] \times [3,10] \times [50,100] \times [10,20] \times [100,200] \times [15,35]$.
We use Gaussian process regression as a machine learning model.
The hyperparameter of the regularization constant $\lambda $ is $0.1$.

\subsubsection{Result}

The execution time of model bridging is 1.1 [s] in the presented computational environment, while simulator calibration requires about 1.3 [h] for $L+1$-th dataset.
The simulator calibration requires $m \times n$ execution of simulation, and each simulation takes 3 [s] in this case.
The results of the mean and standard deviation of the estimated parameters for one-leave-out cross-validation are shown in the bottom rows in Table~\ref{table:ACME_parameters}.
All parameters of estimation by the model-bridging framework are accurate within the standard deviation for $\theta ^{(0)}$ and $\theta ^{(1)}$, respectively.
We can see the effectiveness of a high-dimensional parameter space with a realistic experiment.
From the estimated result of simulation parameters, we can understand that the difference in $\theta_5$ results in different predictions for each situation, while other parameters are constant.
This insight obtained from the interpretable parameters leads to improvements in the production process.

\subsection{Experiment with Simulator for Fluid Dynamics}

\subsubsection{Setting}

\begin{figure}[t]
  \centering
  \includegraphics[width=9.8cm, bb=0 0 685 311]{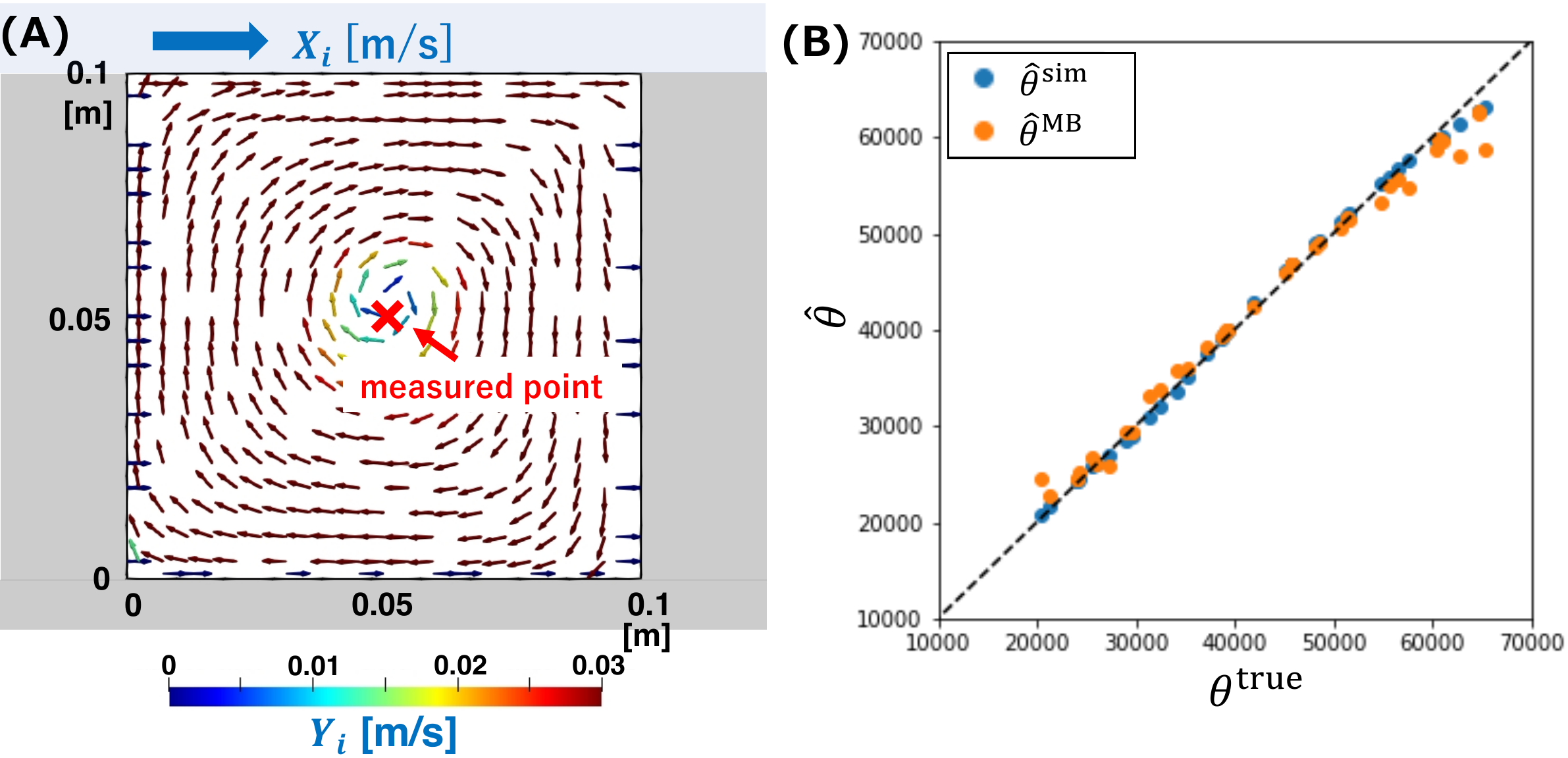}
  \caption{
(A) The experiment of ``cavity,'' which is a two-dimensional square space surrounded by walls (gray) on three sides while moving material (light blue) is located on top of the space.
Input $X_i$ is the velocity of the material on top of the cavity, while output $Y_i$ is the velocity at the point depicted by the x-mark at a specific time.
(B) The estimated result of Reynolds number by simulator calibration ($\hat{\theta}^{\rm sim}$) and model bridging ($\hat{\theta}^{\rm MB}$) as a function of true $\theta$ ($\theta^{\rm true}$).
  }
  \label{fig:cavity_experiment}
\end{figure}

Through computer-aided engineering (CAE) simulations,
we confirmed that our model-bridging algorithm is applicable to the simulation of fluid-dynamics systems.
We employed the typical benchmark in this field, called ``cavity flow experiment,'' with OpenFOAM\textregistered \footnote[2]{\url{https://www.openfoam.com/}} \footnote[3]{\url{https://www.openfoam.com/documentation/tutorial-guide}} (Fig.~\ref{fig:cavity_experiment}~(A)).
We considered a two-dimensional squared space called ``cavity'' fulfilled with a fluid having an unknown Reynolds number.
The Reynolds number is used to help predict flow patterns and velocities in fluid dynamics.
Turbulent flow is somewhat challenging to predict, even though it is ubiquitous in real-world situations.
In this experiment, input $X_i \in \mathbb{R}$ is the velocity of the material on top of the cavity; the output $Y_i \in \mathbb{R}$ is the velocity at the particular point (see Fig.~\ref{fig:cavity_experiment}~(A)); and parameter $\theta \in \mathbb{R}$ is the Reynolds number (see supplementary material for details).
The number of data $n=50$, the number of samples $m=41$, and the number of dataset $L=41$ are generated by different true $\theta_l (=\theta^{\rm true}_l)$.
The prior distribution is defined as the uniform distribution over $[20000,65000]$.
We used Gaussian process regression as a machine learning model.
The hyperparameter of regularization is $\lambda = 1.0^{-5}$.

\subsubsection{Result}

The execution time of model bridging is 2.6 [s] in the presented computational environment, while the simulator calibration requires about 9 [h].
Each simulation takes about 17 [s] in this case.
Figure~\ref{fig:cavity_experiment}~(B) shows the estimated result of $\hat{\theta}^{\rm sim}$ by simulator calibration and $\hat{\theta}^{\rm MB}$ by model bridging as a function of true $\theta$ for $L=41$ dataset with one-leave-out cross-validation.
The dashed line in Fig.~\ref{fig:cavity_experiment}~(B) shows $\theta^{\rm true} = \hat{\theta}^{\rm MB} (=\hat{\theta}^{\rm sim})$ to ensure that the estimation is accurate if the result is on the dashed line.
We can see a reasonable estimation of $\hat{\theta}^{\rm MB}$.
The result of the velocity prediction of velocity $Y_i$ is also reasonably accurate (see the supplementary material).
Human experts can understand why the Reynolds number causes such flow of fluid.

\section{Discussion}
\label{sec:Discussion}

There are many possible options to be discussed in the proposed framework for the individual-use case.
In this study, we assume the given observed dataset as the problem setting. 
Further, there are two other possible ways for problem setting with the assumption of the data generation process: 1) generate data from $f_{\rm ml}(x; \xi) $ and 2) generate data from $f_{\rm sim} (x; \theta) $.
Considering another case with these assumptions of data generation might be meaningful, e.g., when the real observed data are limited or when the simulation has high confidence.
Another option to be discussed is the parametric or non-parametric regression model for the model-bridging function $\hat{T}$.
In this study, we present the practical effectiveness of the model-bridging framework, while a theoretical analysis of the asymptotic behavior of this framework is still desired.

\section{Conclusion}

We propose a novel framework named ``model bridging'' to bridge from the un-interpretable machine learning model to the simulation model with interpretable parameters.
The model-bridging framework enables us to obtain precise predictions from the machine learning model as well as obtain the interpretable simulation parameter simultaneously without the expensive calculations of a simulation.
We confirmed the effectiveness of the model-bridging framework and accuracy of the estimated simulation parameter using production simulation and simulation of fluid dynamics, which are widely used in the real-world manufacturing industry.

%
%
%
%

\newpage

\appendix 

\begin{center}
    {\bf \Large Supplementary Materials}
\end{center}

\section*{A. Explicit Formulation of Norm of Empirical Kernel Mean}

We present the explicit formulation of $\| \hat{\mu}^{\rm MB}_{L+1} - \hat{\mu}^{\rm sim}_{L+1} \| ^2 _{\mathcal{H}}$.
The key for the calculation is the relation $\langle \hat{\mu}, \hat{\mu}^{\prime} \rangle = \sum \sum k_{\theta}(\theta, \theta^{\prime})$ for $\theta$ kernel.
The RKHS norm between estimated $\hat{\mu}^{\rm MB}_{L+1}$ and target $\hat{\mu}^{\rm sim}_{L+1}$ is described as follows:
\begin{eqnarray}
 \| \hat{\mu}^{\rm MB}_{L+1} - \hat{\mu}^{\rm sim}_{L+1} \| ^2 _{\mathcal{H}} 
& = & 2 \left\{ 1 - \langle \hat{\mu}^{\rm MB}_{L+1} , \hat{\mu}^{\rm sim}_{L+1} \rangle \right\} \nonumber \\
& = & 2 \left\{ 1- \sum_{l=1}^L \sum_{j=1}^m \sum_{j^{\prime}=1}^m v_{l} w_{l,j} w_{L+1, j^{\prime}} k_{\theta}(\theta_{L+1, j^{\prime}}, \theta_{L+1, j}) \right\}. \nonumber
\end{eqnarray}
Then, the norm of the empirical kernel mean is obtained.

\section*{B. Detailed Setting of Experiment for Simple Production Simulation}

A production simulation is a general-purpose simulation software package for discrete and interconnected systems, which is used to model various processes such as production, logistics, transportation, and office work.
All processes are implemented in WITNESS, which is a general-purpose simulation software package for discrete and interconnected systems.
The purpose of the production simulation in this experiment is to predict the total production time when the number of products to be manufactured is set.
Figure~\ref{fig:simpleassembly_experiment} shows a typical assembly process for one product with four parts used in this experiment.
The product consists of a ``TOPS'' part, a ``BOTTOMS'' part, and two ``SCREWS.''
The products assembled in the ``ASSEMBLY'' machine are inspected by the ``INSPECTION'' machine before shipping.
The ``INSPECTION'' machine starts when four assembled products arrive, and it can inspect four assembled products simultaneously.
The parameters $\theta_1$ and $\theta_2$ represent the mean and standard deviation in a normal distribution of elapsed time for the ``ASSEMBLY'' machine, respectively.
The parameters $\theta_3$ and $\theta_4$ represent the mean and standard deviation in a normal distribution of elapsed time for the ``INSPECTION'' machine, respectively.

\section*{C. Details of the Experiment with Realistic Production Simulator}

\begin{figure}[t]
  \centering
  \includegraphics[width=11.cm, bb=0 0 720 240]{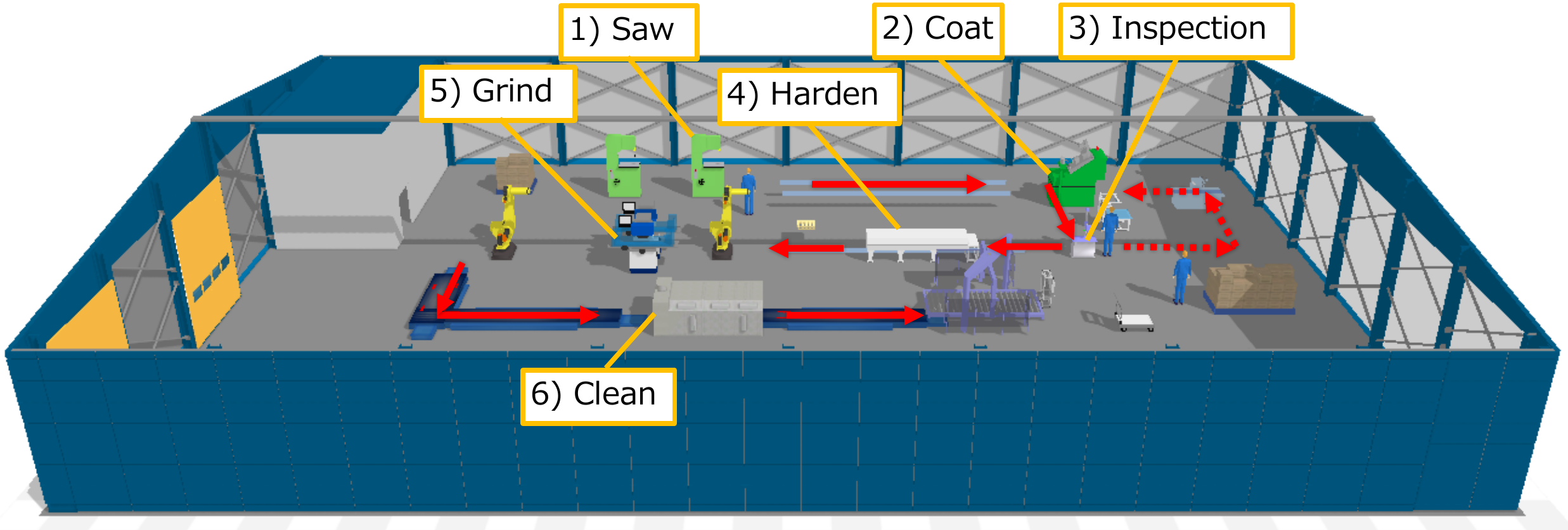}
  \caption{Metal-processing factory manufacturing valves.}
  \label{fig:ACME_overview_3D}
\end{figure}

\begin{figure}[t]
  \centering
  \includegraphics[width=7.cm, bb=0 0 336 220]{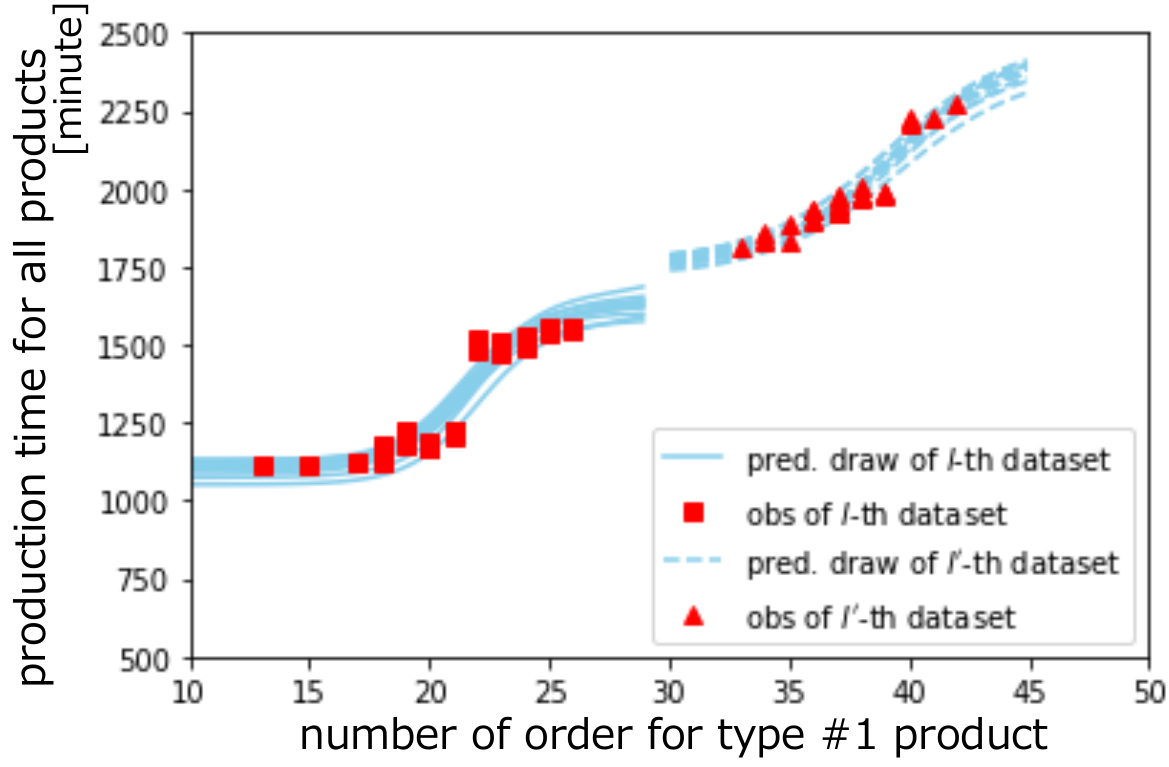}
  \caption{
  As representatives for all dataset, two test datasets are shown. $l$-th dataset denoted as square markers and $l^{\prime}$-th dataset denoted as triangle markers.
The relation between the total production time as $Y_i$ as a function of the number of the units $X_{i, 1}$ to be manufactured as a representative of input vector $X_i = (X_{i, 1}, X_{i, 2}, X_{i, 3})$ as three types of products.}
  \label{fig:ACME_x_y}
\end{figure}

As a realistic experimental setting for a factory manufacturing valves, the process details are described below.
All processes are implemented in WITNESS, which is a general-purpose simulation software package for discrete and interconnection systems.
Figure~\ref{fig:ACME_overview_3D} is an illustration of the simulation model for a realistic experiment.
We defined the total production time as $Y_i \in \mathbb{R}$ when the number of units $X_i \in \mathbb{R}^3$ for three types of products to be manufactured.
Each type of product has different elapsed times at the ``SAW'' process.
Figure~\ref{fig:ACME_x_y} shows the two datasets and regression results by BNN as representatives for all datasets.
The discrete data is originated from a series of batch processing in the simulation model.

{\bf Cutting process: }
The first phase of the manufacturing process begins with the arrival of a pipe having the same diameter and length, 30 cm.
The pipes arrive at a fixed time interval based on the vendor's supply schedule.
Subsequently, each pipe is cut to 10-cm sections along the length.
Thus, three parts can be obtained from one pipe.
For the cutting process, a worker who performs changeover, repair, and disconnection is assigned.
The worker goes for a lunch break once every eight hours.
Thereafter, the parts are transferred from the cutting process to the coating process on a conveyor belt.

{\bf Coating process: }
The cut parts are coated for protection.
In the coating machine, six parts are batch-processed at once.
The coating material must be prepared in the coating machine prior to the part.
Otherwise, the parts will be degraded by the heat.
When the parts ride on the belt conveyor, the sensor detects them, and the coating material is prepared.

{\bf Inspection process:}
After being coated, each part is placed in the inspection waiting buffer before the inspection step.
The inspector will remove the parts individually from the waiting buffer and inspect the coating quality.
If the part fails the quality inspection, the inspector places the part in the recoating waiting buffer.
The coating machine must process the parts of the recoating buffer preferentially.
When the part passes the quality inspection, the inspector sends the part to the curing step.

{\bf Harden process: }
In the harden (quenching) process, up to 10 parts are processed simultaneously on a first-come first-out basis, and each part is quenched for at least one hour.

{\bf Grind process: }
The quenched parts are polished for satisfying the customer's specifications.
Two polishing machines with the same priority are available. Each machine uses special jigs to process four parts simultaneously.
Each of the two polishing machines produces two different types of valves.
Further, 10 jigs exist in the system, and when not in use, they are placed in the jig storage buffer.
A loader fixes the four parts with a jig and sends it to the polishing machine.
The polishing machine sends the jig and four parts to the unloader after the polishing is finished.
The unloader sends the finished parts to the valve storage area and the jig to the jig return area.
The two types of valves are separated and placed in a dedicated valve storage buffer.
As the jig needs to be used again, it is returned from the jig return conveyor to the jig storage buffer.

{\bf Cleaning process: }
The valves removed from the valve storage area are cleaned before shipment.
In the washing machine, five stations are available where the valves can be placed one at a time, and the valves are cleaned in these stations.
Up to 10 valves of each type can be washed simultaneously.
When the valve type is changed, the cleaning head must be replaced.

\section*{D. Details of the Experiment with Fluid-Dynamics Simulator}

\begin{figure}[t]
  \centering
  \includegraphics[width= 7.cm, bb=0 0 511 538]{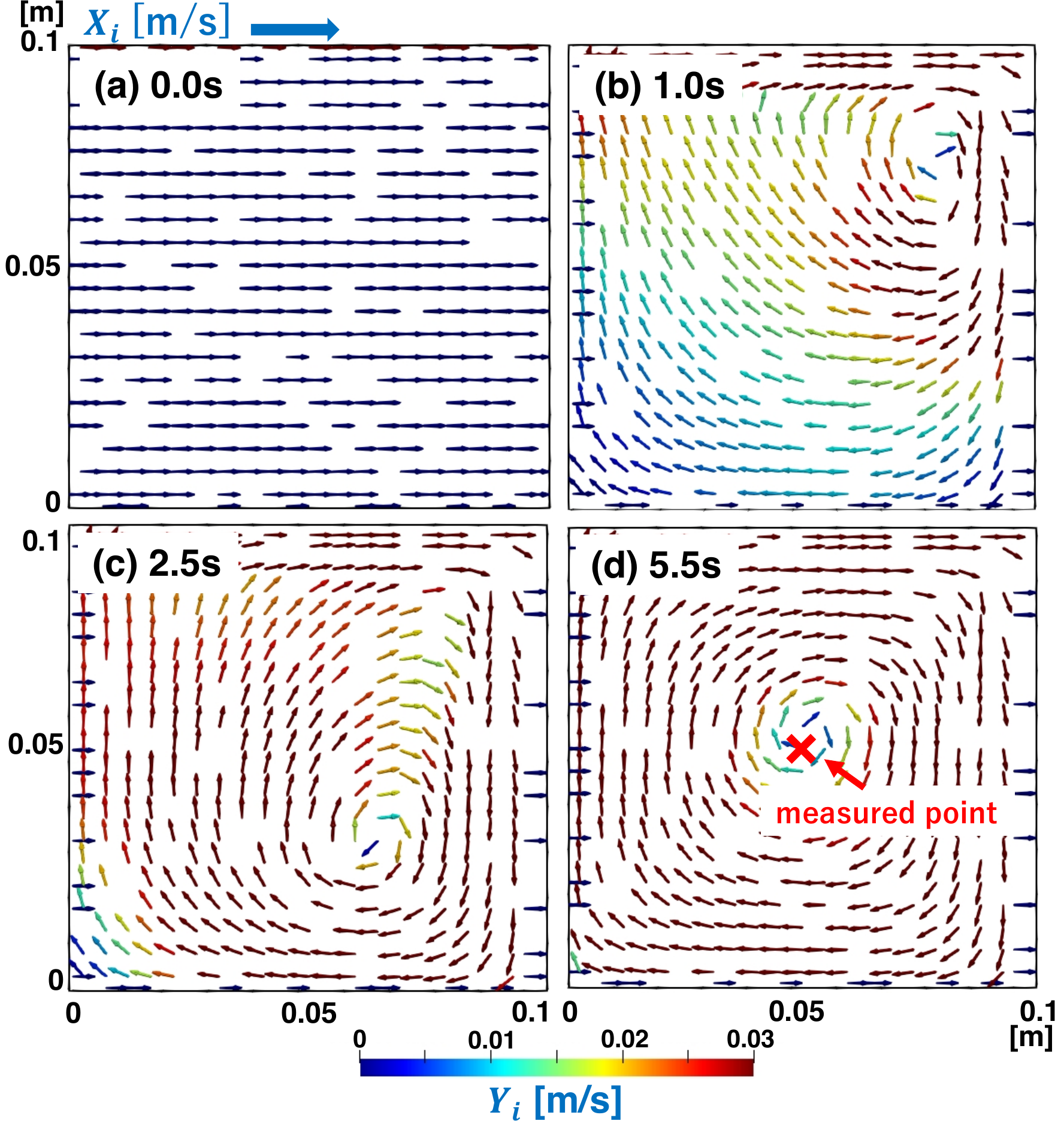}
  \caption{Experiment for ``cavity flow'', which is a typical benchmark in the simulation field of fluid dynamics.
  The Reynolds number condition in this experiment has the turbulent-flow characteristic for observing the time-series data.
In part (a) shows the state at $t = 0$ [s], (b) at $t = 1$ [s], (c) at $t = 2.5$ [s], and (d) at $t = 5.5$ [s].
Here, the fluid velocity at the point depicted by the x-mark at $t = 5.5$ [s] was used for (pseudo-)observation or simulation results.}
  \label{fig:cavity_4figs}
\end{figure}

These experiments are performed using the general-purpose open CAE simulators using the Finite Element Method (FEM) solver; OpenFOAM\textregistered.
OpenFOAM~\textregistered~includes some realistic problems as a tutorial.
The example of ``cavity flow'' is one of the typical benchmarks.
Figure~\ref{fig:cavity_4figs} shows illustrations of the experiment of ``cavity flow'' that are simulated in this study.
Most of the initial experimental settings, such as the number of FEM meshes, cavity size, and so on, were the same as in the tutorial.
The range of the Reynolds number (Re) used was $10000< {\rm Re} < 70000$.
These settings cause the typical turbulent flow, as shown in Fig.~\ref{fig:cavity_4figs}.

\begin{figure}[t]
  \centering
  \includegraphics[width= 7.cm, bb=0 0 395 259]{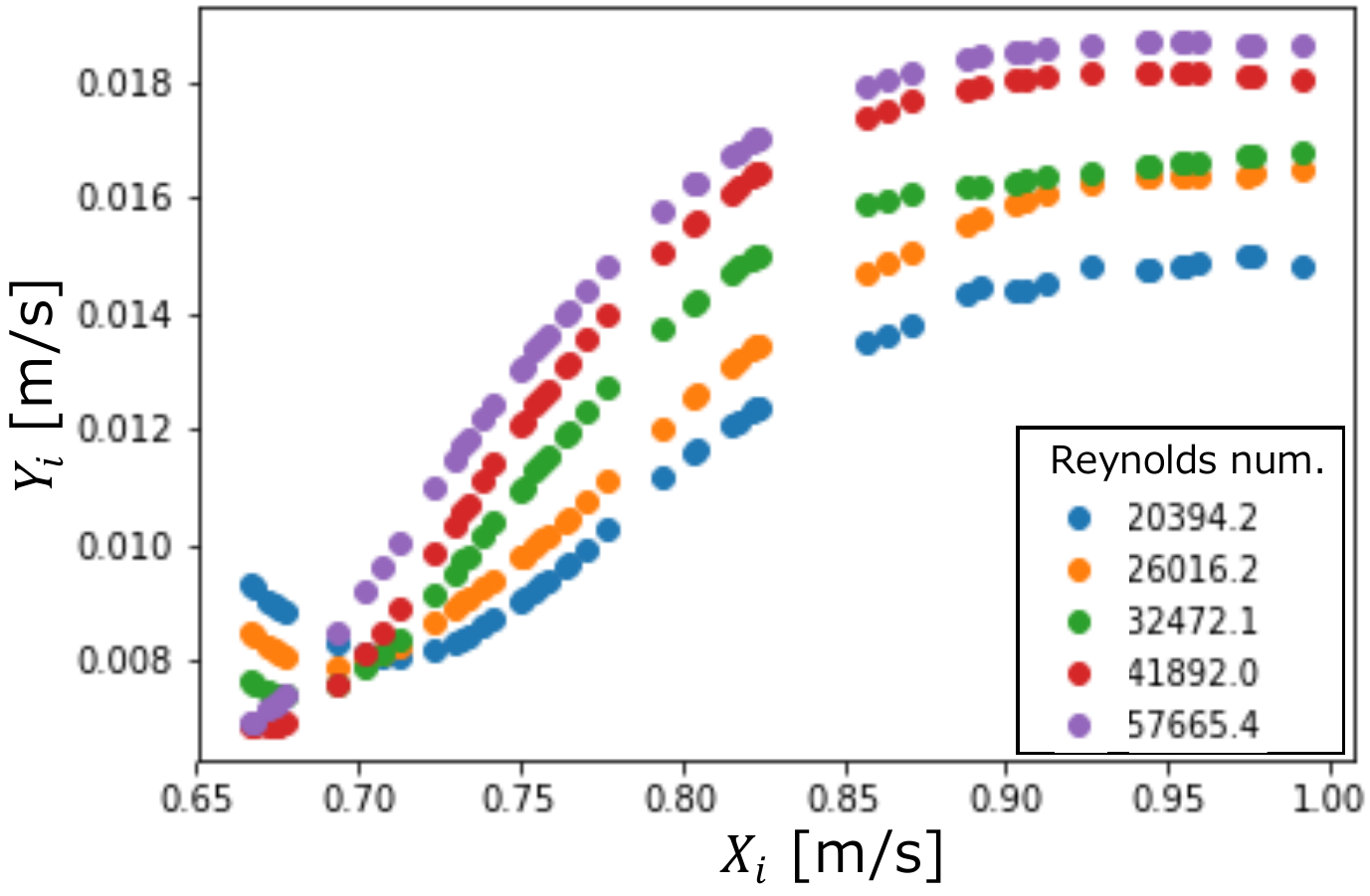}
  \caption{As representatives for all datasets, the relation between $X_i$ and $Y_i$ for five datasets with different parameter $\theta_l (l=1, ...,5)$, which are randomly selected, is shown. $X_i$ is velocity of the material on top of the cavity; $Y_i$ is velocity at the particular point; and parameter $\theta$ is the Reynolds number. }
  \label{fig:cavity_x_y}
\end{figure}

Time-series data are observed where the center of flow is moving, as shown in Fig.~\ref{fig:cavity_4figs}~(a) to (d).
Part (a) of Fig.~\ref{fig:cavity_4figs} shows the state at $t = 0$ [s] (initial state), (b) at $t = 1.0$ [s], (c) at $t = 2.5$ [s], and (d) at $t = 5.5$ [s].
In this experiment, $X_i \in \mathbb{R}$ is velocity of the material on top of the cavity; $Y_i \in \mathbb{R}$ is the velocity at the particular x-marked point at $t = 5.5$ [s] (Fig.~\ref{fig:cavity_4figs}~(d)); and parameter $\theta \in \mathbb{R}$ is the Reynolds number.
Figure~\ref{fig:cavity_x_y} shows the relation between $X_i$ and $Y_i$ for five different dataset as representatives.
It is difficult to select the parametric statistical model as a regression function owing to the non-trivial relation.

\end{document}